\pdfoutput=1

\documentclass[11pt]{article}

\usepackage{acl}

\usepackage{times}
\usepackage{latexsym}
\usepackage{booktabs}
\usepackage{amsmath}
\usepackage{amsfonts}
\usepackage{amssymb}
\usepackage{float}
\usepackage{graphicx}
\usepackage{multirow}
\usepackage{subcaption}
\usepackage{pifont}

\usepackage[T1]{fontenc}

\usepackage[utf8]{inputenc}

\usepackage{microtype}

\definecolor{mygray}{gray}{.9}
\newcommand{\cmark}{\ding{51}}%
\newcommand{\xmark}{\ding{55}}%

\title{DiactTOD: Learning Generalizable Latent Dialogue Acts for Controllable Task-Oriented Dialogue Systems}

\DeclareMathOperator*{\argmin}{arg\,min}

\author{
    Qingyang Wu$^{1}$\Thanks{~Work performed during an internship at AWS AI Labs.}\quad James Gung$^{2}$\thanks{~~Corresponding author.}~\quad Raphael Shu$^{2}$\quad Yi Zhang$^{2}$ \\
    Columbia University$^{1}$\quad AWS AI Labs$^{2}$\\
    \texttt{qw2345@columbia.edu} \\
    \texttt{\{gungj,zhongzhu,yizhngn\}@amazon.com}
}

\begin{document}
\maketitle






\begin{abstract}

Dialogue act annotations are important to improve response generation quality in task-oriented dialogue systems.
However, it can be challenging to use dialogue acts to control response generation in a generalizable way because different datasets and tasks may have incompatible annotations.
While alternative methods that utilize latent action spaces or reinforcement learning do not require explicit annotations, they may lack interpretability or face difficulties defining task-specific rewards.
In this work, we present a novel end-to-end latent dialogue act model (DiactTOD) that represents dialogue acts in a latent space.
DiactTOD, when pre-trained on a large corpus, is able to predict and control dialogue acts to generate controllable responses using these latent representations in a zero-shot fashion.
Our approach demonstrates state-of-the-art performance across a wide range of experimental settings on the MultiWOZ dataset, including zero-shot, few-shot, and full data fine-tuning with both end-to-end and policy optimization configurations.

\end{abstract}

\section{Introduction}

Task-oriented dialogue systems have become increasingly prevalent in recent years, leading to a growth in research on related topics such as dialogue response generation. 
Previous work \cite{DBLP:conf/aaai/YangLQ21,he2022galaxy} found that incorporating dialogue act annotations, representing the illocutionary level of utterances, can enhance the quality of generated responses.
Despite the importance of dialogue act annotations, collecting them can be a time-consuming process that requires human effort. 
Furthermore, existing annotations for dialogue acts are scattered across different datasets and may use different labeling schemes, making it difficult to generalize across tasks. 
As a result, learning to identify and classify general dialogue acts becomes a crucial challenge in the field of task-oriented dialogue systems.


\begin{figure}[t]
    \centering
    \includegraphics[scale=0.6]{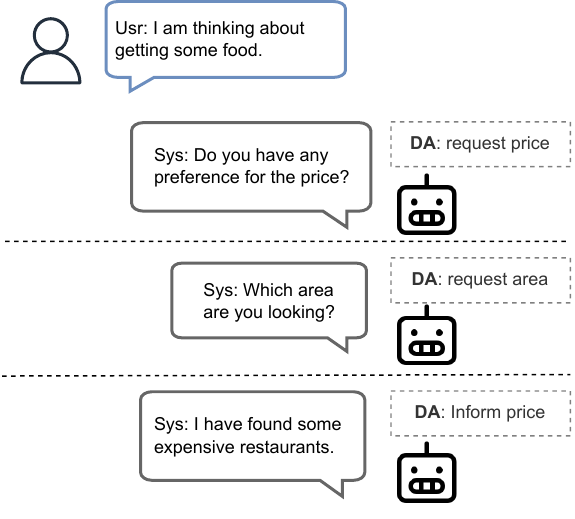}
    \caption{Given different human-readable dialogue acts, the proposed system can produce different responses based on the context.}
    \label{fig:act_example}
\end{figure}

Dialogue acts refer to the underlying intention or purpose of a response in a conversation. 
For example, in Figure~\ref{fig:act_example}, a response might be intended to ask about price or area preference or provide information given the same context.
In task-oriented dialogue systems, it can be useful to classify the dialogue acts of responses in order to generate more appropriate and relevant responses.
One way \cite{chen2013dialogue} to improve the quality of generated responses is to use a dialogue policy model to select the most appropriate dialogue act for a given context. 
However, this approach can be limited in complex or varied situations and may not work well across different datasets. Instead, more advanced techniques may be needed to generate high-quality responses in a wide range of contexts.

An alternative way is to discard predefined semantic dialogue acts and instead use latent action spaces to optimize response generation.
By using latent action spaces, it is possible to generate responses that are more flexible and adaptable to a wider range of situations, without requiring human experts to define the action spaces in advance.
LaRL \cite{zhao2019rethinking} first explores the idea of training an agent to discover underlying patterns and structures in a conversation dataset and to generate responses based on these patterns.
Later work, such as LAVA \cite{lubis2020lava} and DialogVED \cite{chen-etal-2022-dialogved}, extended this idea by using a variational autoencoder (VAE) to improve the performance of the latent action model.
Other approaches, such as PLATO \cite{bao-etal-2020-plato}, have explored using latent action spaces to optimize dialogue agents with large-scale pre-training.

\begin{figure}[t]
    \centering
    \includegraphics[width=0.48\textwidth]{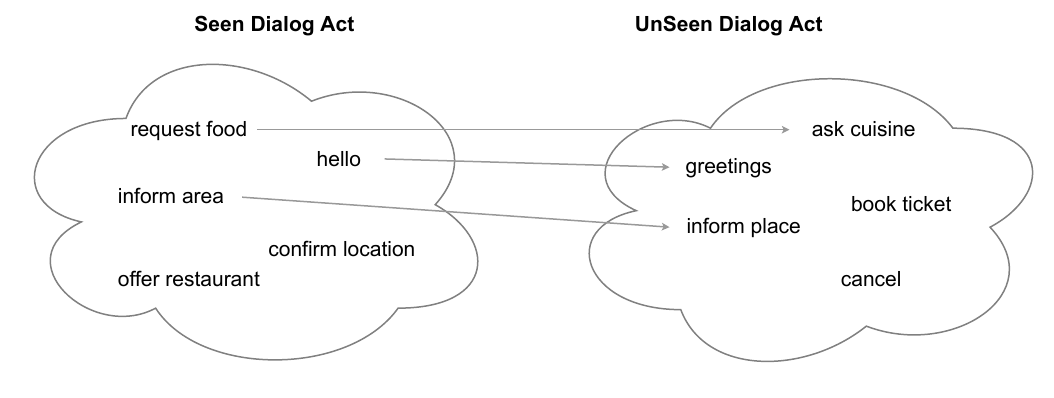}
    \caption{Different datasets have different dialogue act annotation labelsets. How to generalize to unseen dialogue acts becomes a challenge.}
    \label{fig:unseen_act}
\end{figure}

While previous work \cite{zhao2019rethinking,lubis2020lava,chen-etal-2022-dialogved,bao-etal-2020-plato} explored the use of latent action spaces and reinforcement learning for dialogue systems, it has not addressed the possibility of learning general dialogue acts that can be applied across multiple datasets. 
This is an important consideration for task-oriented dialogue systems, which often need to handle a wide range of different tasks and contexts.
In Figure~\ref{fig:unseen_act}, we show examples of the fact that different datasets often have incompatible or inconsistent definitions for dialogue act annotations.
Another limitation of previous approaches is that they fully avoid semantic dialogue act annotations, which can lack controllability and interpretability for the learned actions. This can make it difficult to understand why the system is generating certain responses or to modify its behavior in specific situations. 
As a result, there is a need for new approaches that can learn general dialogue acts across datasets and that provide more control and interpretability for the learned actions.

In this work, we propose a novel method for learning generalized latent dialogue acts that can be applied to new domains for task-oriented dialogues. 
Our method uses sentence-BERT \cite{reimers-2019-sentence-bert} to encode seen dialogue acts into latent representations and a separate policy model to handle context and database information.
To integrate these two components into a single end-to-end model, we modify a pre-trained encoder-decoder model \cite{DBLP:journals/corr/abs-1910-10683,DBLP:conf/acl/LewisLGGMLSZ20} to include the policy model, and further train it to select the best latent dialogue act for a given context.

Our model is designed to perform zero-shot and controllable dialogue response generation, meaning that it can generate appropriate responses without requiring any additional training data. 
To achieve this, we pre-train our model on a large corpus of dialogues and act annotations.
Before pre-training, we fine-tune another model, TANL \cite{DBLP:conf/iclr/PaoliniAKMAASXS21}, with SGD's slot definitions \cite{rastogi2020towards} from a separate dataset to delexicalize the pre-training data to improve its zero-shot capability. 
These steps allow our model to learn generalizable latent dialogue act representations and generate appropriate responses that can be applied to new tasks and datasets without additional fine-tuning. 

We evaluate the effectiveness of our model on the MultiWOZ  \cite{DBLP:conf/emnlp/BudzianowskiWTC18} dataset, a widely-used benchmark for task-oriented dialogue generation.
During inference, we control the dialogue acts using the provided schema and targeted objective to generate better system responses.
We test our model in a range of experimental settings, including zero-shot, few-shot, and full fine-tuning response generation for both end-to-end and policy optimization configurations.
In all of these settings, our model outperforms previous baselines and achieves state-of-the-art performance.


Our main contributions in this work can be summarized as follows: 
\begin{itemize}
\item We present a novel end-to-end latent dialogue act model that represents arbitrary dialogue acts in latent space and can predict and control these acts to generate better responses.

\item We pre-train our model with a semi-supervised method for learning latent dialogue acts that can generalize across different datasets with different act labels. 

\item Our model DiactTOD achieves state-of-the-art performance on the MultiWOZ dataset in a range of experimental settings, including zero-shot, few-shot, and full fine-tuning in both end-to-end and policy optimization configurations.

\end{itemize}











\section{Related Work}



Response generation is an important task in task-oriented dialogue systems. There have been many previous approaches \cite{HosseiniAsl2020ASL,DBLP:conf/eacl/WuZLY21,DBLP:conf/acl/GuWWSY20,DBLP:conf/acl/SuSMG0LZ22,he2022galaxy,https://doi.org/10.48550/arxiv.2211.16773,DBLP:journals/corr/abs-2210-08917,DBLP:journals/corr/abs-2305-13710} proposed to improve the task-oriented dialogue systems.
One direction is the use of dialogue act annotations to improve the quality of responses in task-oriented dialogue systems. 
For example, SimpleTOD \cite{HosseiniAsl2020ASL} and UBAR \cite{DBLP:conf/aaai/YangLQ21} generate dialogue acts as part of the response generation process. 
PPTOD \cite{DBLP:conf/acl/SuSMG0LZ22} uses the context as a prompt and dialogue act generation for multi-task learning. 
Recently, GALAXY \cite{he2022galaxy} proposed a method that uses pre-training on a large corpus of dialogues with dialogue act annotations as an auxiliary objective to improve the quality of the generated responses.
However, these methods are limited by the fact that different datasets may have incompatible or inconsistent dialogue act annotations for learning generalizable representations.
To address this problem, previous work \cite{he2022galaxy,DBLP:conf/interspeech/PaulGH19} has attempted to define a new universal schema for dialogue acts. 
However, these approaches are either overly simplified or require additional human annotations, limiting their effectiveness and practicality.


In addition to using explicit annotations of dialogue acts, researchers have also explored alternative methods to improve response generation, such as using latent action spaces and implementing reinforcement learning techniques.
These approaches aim to improve the overall task success rate of generated responses.
LaRL \cite{zhao2019rethinking} uses latent dialogue acts trained with reinforcement learning instead of surface-form dialogue acts to control response generation which results in the best task score. 
LAVA \cite{lubis2020lava} further improves over LaRL by utilizing a variational autoencoder (VAE) to learn an informed and semantic prior when optimizing the latent action spaces, achieving state-of-the-art Success and Inform scores on MultiWOZ.
KRLS \cite{https://doi.org/10.48550/arxiv.2211.16773} is another recent approach that applies reinforcement learning to pre-trained language models.
This approach utilizes a specifically designed objective function that focuses on learning the keywords in the input, with the goal of improving the overall performance of the language model.
In our work, we adopt a similar approach but use dialogue act annotations to assign semantic meanings to the latent representations, allowing the model to learn generalizable and controllable latent dialogue acts, which improves the quality of generated response.


Pre-training with a large corpus of dialogues has been a widely adopted technique to enhance the response generation quality in dialogue systems  \cite{DBLP:conf/acl/ZhangSGCBGGLD20,DBLP:conf/eacl/RollerDGJWLXOSB21}. 
In the context of task-oriented dialogue systems, several recently proposed approaches have demonstrated the effectiveness of pre-training.
GALAXY \cite{he2022galaxy} pre-trains the model with a collection of dialogue datasets with dialogue act annotations.
GODEL \cite{Peng2022GODELLP} uses a larger dataset and model size, and it also incorporates the grounding of database results in the context.
This allows it to achieve good performance under few-shot settings on the MultiWOZ dataset.
In contrast, our work uses a smaller set of pre-training datasets but with more robust data processing techniques.
We use the complete dialogue acts in each dataset without any simplification. 
We also train another model TANL \cite{DBLP:conf/iclr/PaoliniAKMAASXS21} to delexicalize the pre-training data to improve the model's zero-shot and few-shot capabilities. 

\begin{figure*}[t]
    \centering
    \includegraphics[width=0.95\textwidth]{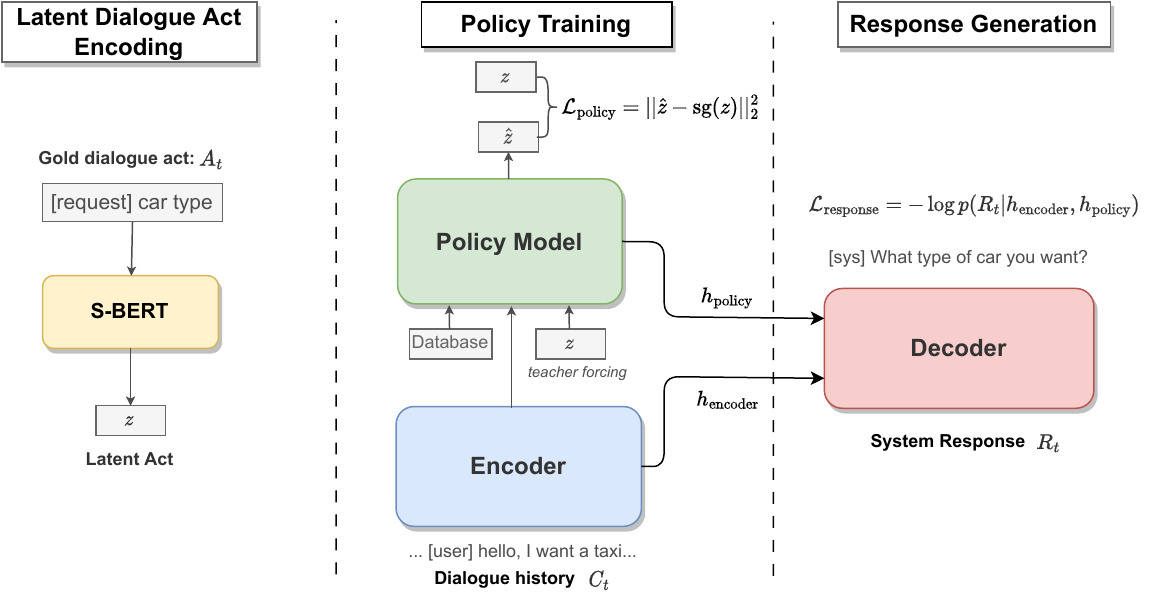}
    \caption{Overview of the training pipeline, which includes three stages: latent dialogue act encoding, policy training, and response generation.
    During training, dialogue acts are first encoded into latent vectors and then passed to a policy model to control the final response generation.}
    \label{fig:overview}
\end{figure*}

\section{DiactTOD Approach}

In this section, we first provide a brief overview of the traditional end-to-end task-oriented dialogue systems. 
Then, we delve into the specifics of how our proposed latent dialogue act model operates, by providing details on both its training and inference processes, which offers a new approach to modeling dialogue acts.
Finally, we discuss how this model can be used to control response generation for a more efficient and accurate dialogue system.




\subsection{End-to-End Task-Oriented Dialogue}

An end-to-end task-oriented dialogue system generates a system response $R_t$ at turn $t$ based on the dialogue history context $C_t$ and the database result $DB_t$.
The history context $C_t$ contains the previous user utterances $U_{1:t}$ and the system responses $R_{1:t}$. 
To get the database search result $\text{DB}_t$, a dialogue state tracking (DST) model would need to output the belief state $B_t$.
To leverage the dialogue act annotations, the model also generates act $A_t$ for dialogue policy learning.
This allows the model to effectively guide the conversation and produce accurate and appropriate responses.
\begin{align}
    \mathcal{L}_{\text{act}} &= - \log p(A_t | C_t, \text{DB}_t)
\end{align}

The final system response is generated conditional to the history context $C_t$, the database result $\text{DB}_t$, and the dialogue act $A_t$.
\begin{align}
    \mathcal{L}_{\text{response}} &= - \log p(R_t | C_t, \text{DB}_t, A_t)
\end{align}

In practice, the dialogue acts $A_t$ and the system response $R_t$ are concatenated during the training and generation process to improve the decoder's performance. 
However, the surface form of dialogue acts has limitations in terms of generalization, as different datasets and tasks may have different formats for representing dialogue acts. This can make it difficult to apply the model to different settings.














\subsection{Generalizable Latent Dialogue Act}

Figure~\ref{fig:overview} shows the overview of our approach. 
We divide the pipeline into three parts: latent dialogue act encoding, policy training, and response generation.

\textbf{Latent dialogue act encoding:} To overcome the generalization issues associated with the surface form of dialogue acts, we use sentence-BERT (S-BERT) to encode the dialogue acts into embeddings and we have:
\begin{align}
     z &= \text{S-BERT} (A_t)
\end{align}

This allows different annotations with the same meaning to have similar representations while leveraging the semantic knowledge contained in the encoder to improve generalization.

\hfill

\textbf{Policy Training:} On top of the encoder-decoder architecture, we have introduced a policy model that serves as a way to learn the dialogue policy. 
This model operates similarly to the decoder in an autoregressive manner.
It takes in the database search result $\text{DB}_t$ and the encoder's hidden states $h_\text{encoder}$ as input, and produces a predicted latent dialogue act vector $\hat{z}$ that is optimized to closely match the true latent dialogue act vector $z$.
We use the mean squared error (MSE) loss function to minimize their distance:
\begin{align}
     \hat{z} &= \text{Policy} ( \text{DB}_t, h_\text{encoder}) \\
  \mathcal{L}_{\text{policy}} &= || \hat{z}  - \text{sg}(z) ||_2^2
\end{align}
where $sg$ means stop gradient. This increases the stability of the training. 
During training, the policy model is trained using a technique called teacher forcing, where the true latent dialogue act vector $z$ is provided as input to the model. 
To ensure that the model does not leak any ground truth dialogue act information, a unidirectional attention mask is used.

Then, the true latent dialogue act vector $z$ is fed into the policy model with teacher forcing to produce the policy model's hidden state:
\begin{align}
     h_\text{policy} &= \text{Policy} ( \text{DB}_t, h_\text{encoder}, z)
\end{align}

\textbf{Response generation:} The final system response is generated by the decoder, which takes both the hidden states of the encoder $h_\text{encoder}$ and the hidden states of the policy model $h_\text{policy}$ as the input.
\begin{align}
    h_\text{encoder} &= \text{Encoder} ( C_t) \\
    \mathcal{L}_{\text{response}} &= - \log p(R_t | h_\text{encoder}, h_\text{policy})
\end{align}

This allows the decoder to generate appropriate responses while enabling controllability with the policy model, as the decoder can take into account the dialogue context and the predicted latent dialogue act. 

The final training loss is defined as the sum of the policy loss and the response loss: 
\begin{align}
    \mathcal{L}_\text{training} &= \alpha \mathcal{L}_{\text{policy}} + (1- \alpha)\mathcal{L}_{\text{response}} 
\end{align}
where $\alpha$ is a hyperparameter to balance the magnitude of losses.

\begin{figure}[t]
    \centering
    \includegraphics[width=0.45\textwidth]{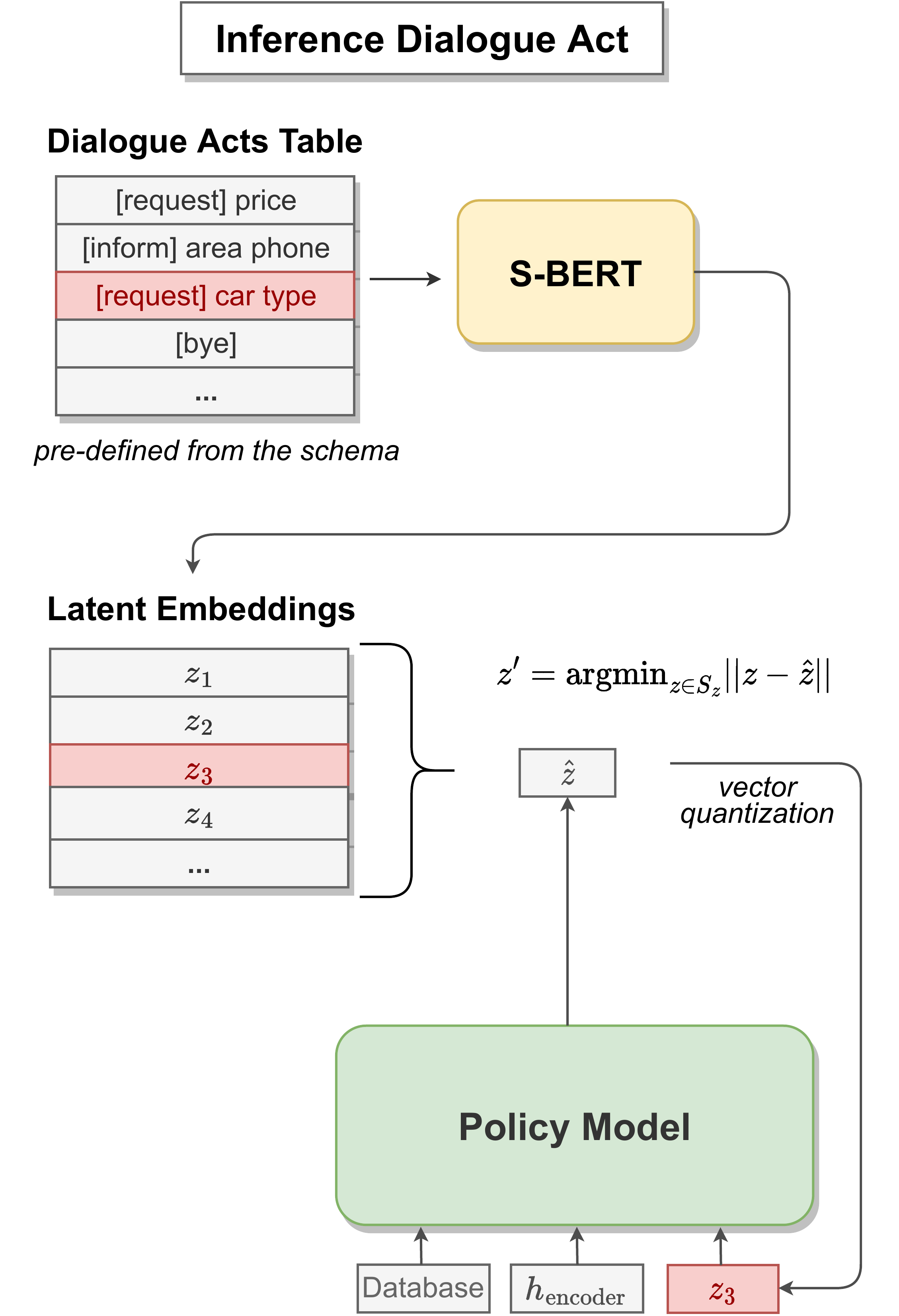}
    \caption{During inference, we select the closest dialogue act based the predicted dialogue act. Note that the set of valid dialogue acts can be filtered based on the task or context.}
    \label{fig:inference}
\end{figure}
\hfill

\textbf{Inference:} During the inference phase (depicted in Figure~\ref{fig:inference}), we pre-define a table $S_z$ that includes all possible combinations of dialogue acts. 
This allows us to create a set of embeddings for the dialogue acts, where each act can be treated as a unique "word" in a specialized vocabulary. 
This table contains all possible combinations of dialogue acts that can be derived from the training dataset. 
Alternatively, if the schema of dialogue acts is known, we can manually construct such a table consisting of valid combinations. 
This can be particularly useful in a zero-shot setting. 
In this scenario, where we do not have a training set for a specific domain, having a set of predefined dialogue acts can allow the model to still generate semantically valid responses without any training.

Once the predicted latent dialogue act vector $\hat{z}$ is generated, it is used to retrieve the most appropriate latent dialogue act from the embedding table $S_z$. 
This is done by using a technique called vector quantization, which allows us to select the latent dialogue act that is closest to the predicted vector. 
This helps reduce the representation mismatch of the predicted latent dialogue between training and inference.
\begin{equation}
    z' = \argmin_{z \in S_z} ||z - \hat{z}||
\end{equation}

After the closest latent dialogue act is retrieved from the embedding table using vector quantization, it is fed back into the policy model. 
The decoder then generates the final system response by conditioning on both the encoder's hidden states and the policy model's hidden states.

\subsection{Controllable Response Generation}

The policy model uses a pre-processed embedding table to predict dialogue acts.
By filtering the embedding table to include only relevant dialogue acts, we can control the predicted dialogue acts during inference. 
This allows the model to focus on generating more appropriate and relevant responses that are tailored to the specific context or task, which improves the overall efficiency and accuracy of the dialogue system. 

For example, if the dialogue act table contains some combinations that lack requesting or informing for certain slots, we can filter these dialogue acts out of the embedding table during inference. 
This helps guide the generation of responses to make more requests or provide more information for those specific slots. This can be particularly useful in scenarios where the user's goal is to obtain specific information or complete a certain task and the model can make more requests or provide more information for the relevant slots. 
In this way, the model can quickly adapt to specific scenarios or domains and respond in a more appropriate and relevant way to the user's needs and goals.

\section{Pre-training Latent Acts}

\begin{table}[h]
    \centering
    \resizebox{0.4\textwidth}{!}{
    \begin{tabular}{c| c c}
    \toprule
    \textbf{Dataset Name} & \textbf{Act Label?} & \textbf{\# Utterances}  \\
    \midrule
    SGD & \cmark & 463,284 \\
    STAR & \cmark & 107,846  \\
    MSRe2e & \cmark & 74,686  \\
    Frames & \cmark & 19,986 \\
    MetaLWOZ & \xmark & 356,268 \\
    \bottomrule
    \end{tabular}
    }
    \caption{Pre-training datasets statistics. For datasets without dialogue act labels, we use system responses as a proxy for the dialogue act.}
    \label{tab:dataset}
\end{table}


To learn generalizable latent dialogue acts and achieve competitive performance on downstream tasks without any additional fine-tuning, our model undergoes pre-training on a selection of task-oriented dialogue datasets shown in Table~\ref{tab:dataset}.
Specifically, we have chosen four datasets that are annotated with dialogue acts and one dataset that does not contain any dialogue act annotations. 
Detailed descriptions of these datasets can be found in the appendix.









To ensure consistency across all datasets for pre-training, we pre-process the datasets with the same tokenization and truncation of dialogues when they exceed a certain length. 
Additionally, we incorporate database search results as an input token to indicate the number of matches.
A large portion of utterances in these datasets do not have dialogue act annotations.
To effectively pre-train on those datasets, we utilize the system response as a proxy for the dialogue act. 
This allows the policy model to generalize to new and unseen dialogue acts. 
Our experiments have shown this approach to be effective.

In task-oriented response generation, system responses are typically in a delexicalized form, which means that specific values of certain variables are replaced by placeholders. 
To enable this automatic delexicalization during response generation, we use the model TANL (Translation between Augmented Natural Languages) \cite{tanl}. 
This model can extract slot spans from the input sentence.
We fine-tune the TANL model with the SGD's predefined slot definitions.
For downstream tasks and evaluation, we ensure compatibility by defining a one-to-one mapping of the SGD's slot definitions with the slots in the MultiWOZ dataset.

\begin{table*}[t]
    \centering
    \resizebox{0.7\textwidth}{!}{
    \begin{tabular}{l |c |c c c c }
    \toprule
    \multirow{2}{*}{\textbf{Model}} & \multirow{2}{*}{\textbf{\# Examples}} & \multicolumn{4}{c}{\textbf{Policy optimization}} \\
    & & \textbf{Inform} & \textbf{Success} & \textbf{BLEU}  & \textbf{Combined} \\
    \midrule
    $\text{DialoGPT}_\text{base \quad}$ & 50 & 38.70 & 3.00 & 0.20  & 21.05 \\
    $\text{DialoGPT}_\text{large}$ & 50  & 62.40 & 34.70 & 10.52 & 59.06 \\
    \midrule
    $\text{T5}_\text{base}$ & 50 & 60.60 & 22.50 & 4.31 & 45.86 \\
    $\text{T5}_\text{large}$ & 50  & 71.50 & 56.20 & 12.69 & 76.54 \\
    \midrule
    $\text{GODEL}_\text{base}$ & 50 & 67.60 & 46.10 & 12.81 & 69.72 \\
    $\text{GODEL}_\text{large}$ & 50 & 81.60 & 62.10 & \textbf{14.07} & 85.90 \\
    $\text{GODEL}_\text{GPT-J}$ & 50  & 60.50 & 21.00 & 6.27 & 47.01 \\
    $\text{GODEL}_\text{GPT-3}$ & 50 & 68.80 & 19.90 & 6.72 & 51.06 \\
    \midrule
    $\text{DiactTOD}$ & 0  & \textbf{93.60} & \textbf{71.40} & 4.20 & \textbf{86.70} \\
    $\text{DiactTOD}$ & 50 & \textbf{94.60} & \textbf{78.90} & 10.75 & \textbf{97.05} \\
    \bottomrule
    \end{tabular}
    }
    \caption{Low-resource experimental results. All experiments are done in the policy optimization setting. For few-shot, we fine-tuned the model with 50 examples.}
    \label{tab:low-resource}
\end{table*}

\begin{table*}[t]
    \centering
    \resizebox{0.9\textwidth}{!}{
    \begin{tabular}{l|c c c c | c c c c}
    \toprule
    \multirow{2}{*}{\textbf{Model}} & \multicolumn{4}{c|}{\textbf{End-to-end}} & \multicolumn{4}{c}{\textbf{Policy optimization}} \\
    & \textbf{Inform} & \textbf{Success}  & \textbf{BLEU}   & \textbf{Combined} & \textbf{Inform} & \textbf{Success}  & \textbf{BLEU}   & \textbf{Combined} \\
    \midrule
    UBAR & 83.4 & 70.3 & 17.6 & 94.4 & - & - & - & - \\
    PPTOD & 83.1 & 72.7 & 18.2 & 96.1 & - & - & - & - \\
    RSTOD & 83.5 & 75.0 & 18.0 & 97.3 & - & - & - & - \\
    BORT & 85.5 & 77.4 & 17.9 & 99.4 & - & - & - & - \\
    MTTOD & 85.9 & 76.5 & 19.0 & 100.2 & - & - & - & - \\
    HDNO & - & - & - & - & 93.3 & 83.4 & 17.8 & 106.1 \\
    GALAXY & 85.4 & 75.7 & 19.6 & 100.2 & 92.7 & 83.5 & \textbf{19.9} & 108.1 \\
    MarCo & - & - & - & - & 94.5 & 87.2 & 17.3 & 108.1 \\
    Mars & 88.9 & 78.0 & \textbf{19.9} & 103.4 & - & - & - & - \\
    KRLS & 89.2 & 80.3 & 19.0 & 103.8 & 93.1 & 83.7 & 19.1 & 107.5 \\
    \midrule
    DiactTOD & \textbf{89.5} & \textbf{84.2} & 17.5 & \textbf{104.4} & \textbf{94.8} & \textbf{90.2} & 17.8 & \textbf{110.3} \\
    \bottomrule
    \end{tabular}
    }
    \caption{MultiWOZ Response generation evaluation. ``-'' means that this setting's performance is not reported. (Combined Score=(Inform + Success)*0.5 + BLEU)}
    \label{tab:full-fine-tune}
\end{table*}

\section{Experiment Setup}

We initialize our model with T5-base and pre-train our model on the previously mentioned datasets.
We evaluate our model on the multi-domain task-oriented dialogue dataset MultiWOZ \cite{DBLP:conf/emnlp/BudzianowskiWTC18}.
It contains 8,438/1,000/1,000 dialogues for training, validation, and testing, respectively.
There are seven different domains, including hotel, hospital, police, restaurant, train, and taxi.
We use MultiWOZ 2.2 \cite{zang-etal-2020-multiwoz} to be compatible with the standardized evaluation script \cite{DBLP:journals/corr/abs-2106-05555}. 
We evaluate our approach under different scenarios, such as zero-shot, few-shot, and fine-tuning with the full dataset, with both end-to-end and policy optimization configurations to evaluate the robustness and flexibility of our model.

We use standardized evaluation metrics\footnote{\url{https://github.com/Tomiinek/MultiWOZ_Evaluation}} with Inform, Success rates, and BLEU scores. 
\textbf{Inform} measures the extent to which the system provides sufficient and relevant information to fulfill the user's information needs.
\textbf{Success} evaluates the performance in completing the user's goal.
Also, we evaluate the model’s zero-shot dialogue act prediction capabilities on an unseen dataset.

To provide a comprehensive evaluation, we separately compare our model's performance against several strong baselines in both low-resource settings and full fine-tuning settings.
In low-resource settings, we compare our model with DialoGPT \cite{DBLP:conf/acl/ZhangSGCBGGLD20}, T5 \cite{DBLP:journals/corr/abs-1910-10683},
and GODEL \cite{Peng2022GODELLP}.
GODEL and DialoGPT are trained with a much larger dialogue corpus.
Those models require a minimum of 50 training examples to adapt to MultiWOZ training data,
while our work can perform zero-shot response generation without any fine-tuning.

For the full dataset fine-tuning settings, we compare with models on the existing leaderboard of MultiWOZ.
We evaluate both end-to-end and policy optimization settings.
This includes UBAR \cite{DBLP:journals/corr/abs-2106-05555}, PPTOD \cite{DBLP:conf/acl/SuSMG0LZ22}, RSTOD \cite{DBLP:journals/corr/abs-2208-07097}, BORT \cite{DBLP:conf/naacl/Sun0W022}, MTTOD \cite{Lee2021ImprovingET}, HDNO \cite{DBLP:journals/corr/abs-2006-06814}, GALAXY \cite{he2022galaxy}, MarCO \cite{DBLP:conf/acl/WangTWQY20}, Mars \cite{DBLP:journals/corr/abs-2210-08917}, and KRLS \cite{https://doi.org/10.48550/arxiv.2211.16773}.
To obtain database search results in the end-to-end setting, we use MTTOD's dialogue state tracker, which is trained jointly during fine-tuning.
We follow previous methods and append the dialogue act in front of the system responses to improve performance.

\section{Experiments}
In this section, we first show the experimental results under the low-resource and full fine-tuning settings.
Next, we analyze the model's zero-shot capability to predict dialogue acts.
Finally, we perform ablation studies for the proposed model to demonstrate the impact of dialogue act control and pre-training data.

\subsection{Low-resource Settings}

Table~\ref{tab:low-resource} shows the performance of our model in low-resource settings.
We evaluate the performance of our model under zero-shot settings and also fine-tune it using 50 randomly selected dialogues, similar to the approach used by the GODEL model.
The experiments here are done in the policy optimization setting.

Our model outperforms the best GODEL model by achieving a higher combined score of $86.70$ without any fine-tuning, and an even higher score of $97.05$ after fine-tuning.
In particular, our model achieves better scores in Inform and Success metrics, indicating that our model is better able to satisfy the users' information needs.
GODEL model has a higher BLEU score, which is likely due to the larger pre-training corpus used to train the model.

\subsection{Full Fine-tuning Settings}

To evaluate the effectiveness of our model in full dataset fine-tuning settings, we conduct experiments with both end-to-end and policy optimization configurations. 
The results, as shown in Table~\ref{tab:full-fine-tune}, demonstrate that our model achieves state-of-the-art performance, with a combined score of $104.4$. 
In particular, our model outperforms the other models in the Inform and Success metrics, indicating that our model is able to provide more relevant and complete information to satisfy the users' information needs.
Our model receives a slightly worse BLEU score. We suspect this is because the resulting responses contain more information relevant to the user request than the ground truth responses.

\begin{table}[t]
    \centering
    \resizebox{0.48\textwidth}{!}{
    \begin{tabular}{l | c c c c}
        \toprule
        \textbf{Settings} & \textbf{Inform} & \textbf{Success} & \textbf{BLEU} & \textbf{Comb.}\\
        \midrule
        full end-to-end  & 89.5 & 84.2 & 17.5 & 104.4 \\
         \quad - pretrain & 87.7 & 78.9 & 19.7 & 103.0 \\
         \quad - control & 84.9 & 76.2 & 19.8 & 100.4 \\
         \, + gold act & 93.0 & 89.6 & 29.6 & 120.8 \\
         \midrule
        zero-shot policy  & 94.6 & 71.4 & 4.2 & 86.7 \\
         \quad - control & 93.8 & 55.4 & 6.6 & 81.2 \\
         \bottomrule
    \end{tabular}
    }
    \caption{Ablation studies for end-to-end full training settings and zero-shot policy optimization settings.}
    \label{tab:ablation}
\end{table}


\subsection{Zero-shot Dialogue Act Prediction}

\begin{figure}[t]
    \centering
    \begin{subfigure}[b]{0.34\textwidth}
    \includegraphics[width=\textwidth]{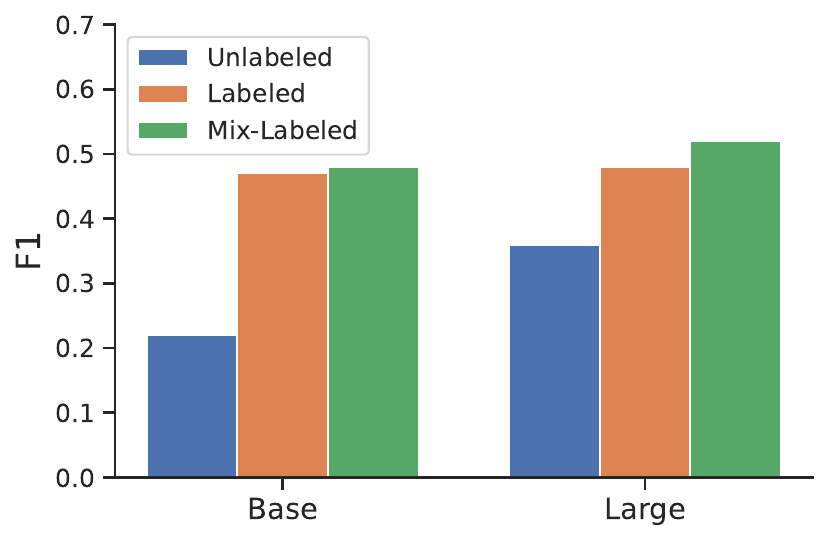}
    \end{subfigure}
    \begin{subfigure}[b]{0.3\textwidth}
    \includegraphics[width=\textwidth]{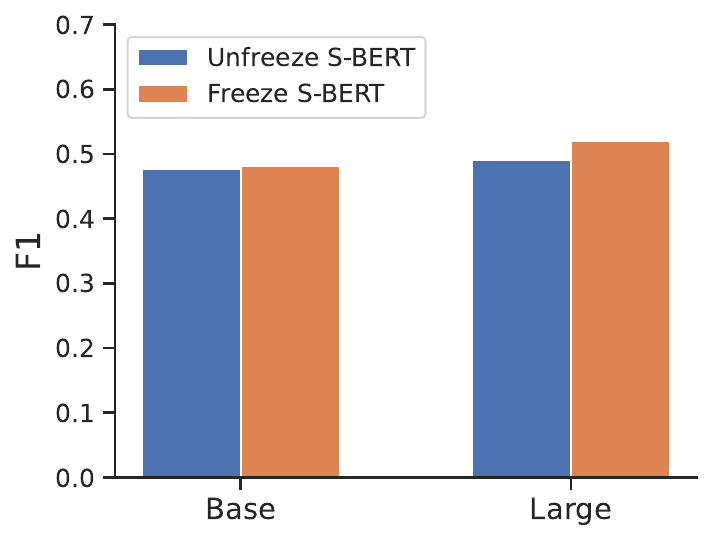}
    \end{subfigure}
    \caption{Zero-shot dialogue act prediction F1 score.}
    \label{fig:act_pred}
\end{figure}

We evaluated the model's capability to predict dialogue acts without any downstream fine-tuning. 
We pre-defined a set of possible dialogue acts by using the dialogue act schema from the training set.
We first tested the effects of different pre-training configurations.
Note that the data is divided into two categories: one with dialogue act labels and one without.
Thus, we evaluated the model pre-trained with unlabeled, labeled, or mix-labeled act annotation data separately.
Additionally, we tested the effect of freezing the sentence-BERT model during training to see its impact on the performance of the overall model.
The results are shown in Figure~\ref{fig:act_pred}.
We observed that pre-training with mixed-label data has the best performance, and freezing the sentence-BERT model had minimal effects on the dialogue act prediction F1.


\begin{figure}[t]
    \centering
    \begin{subfigure}[b]{0.34\textwidth}
    \includegraphics[width=\textwidth]{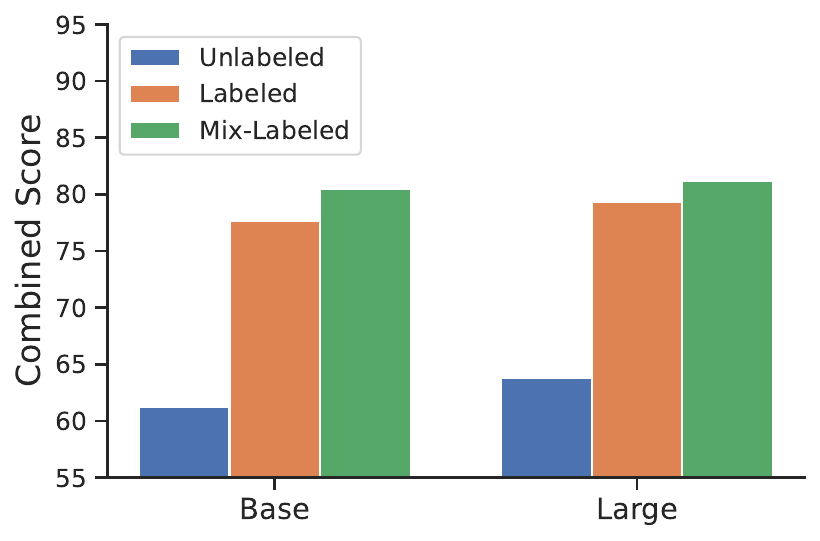}
    \end{subfigure}
    \begin{subfigure}[b]{0.3\textwidth}
    \includegraphics[width=\textwidth]{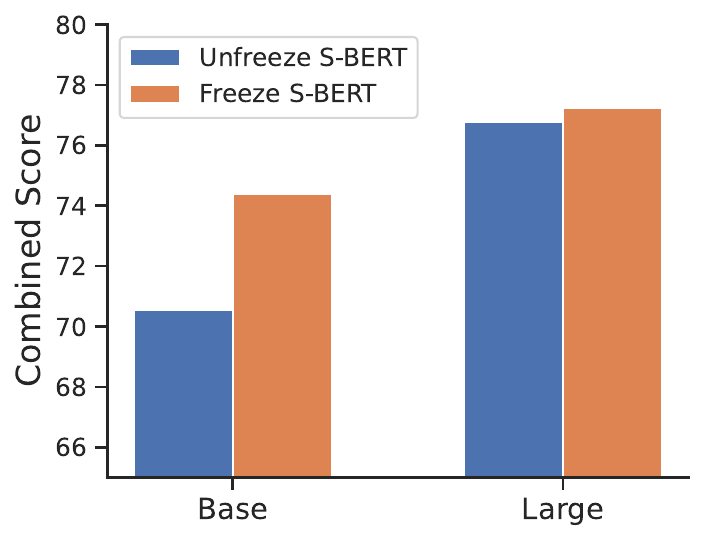}
    \end{subfigure}
    \caption{Response generation combined score.}
    \label{fig:ablations}
\end{figure}

\subsection{Ablations and Analysis}

We conduct similar experiments to the previous section to evaluate the effects of different pre-training configurations.
The experiments here are conducted in the zero-shot setting, without any dialogue act control.
The results are shown in Figure~\ref{fig:ablations}.
Using the labeled data during pre-training significantly improves the performance of the model.
Mixing unlabeled data and labeled data leads to even better performance.
We also observe that for the small model, freezing sentence-BERT during training can significantly improve the performance, but it has less of an effect for the large model.

Then, we evaluated the effects of pre-training and controllable response generation.
The results are shown in Table~\ref{tab:ablation}.
In the full end-to-end fine-tuning setting, we first tested removing pre-training.
From the table, we observed a decrease in the performance of the model for Inform (-2.0\%) and Success (-6.3\%), but an increase in the BLEU score (+10.1\%).
Then, we further tested the model without pre-training and removing the dialogue act response control, allowing the model to predict the dialogue act without any constraints.
It has a combined score of $100.4$, which is close to the reported MTTOD performance, indicating the trade-off when using controlled response generation.
We also tested using gold dialogue acts for our final pre-trained model as a reference for comparison.
In the zero-shot setting, we observed similar patterns when removing dialogue act control, but the performance decrease is more significant. Specifically, the Success rate dropped from $71.4$ to $55.4$, suggesting that our controlled response generation with dialogue acts is more effective in the low-resource setting than in the full fine-tuning setting.

\section{Conclusion}

In this work,  we present a novel end-to-end latent dialogue act model (DiactTOD) that represents dialogue acts in a latent space to improve the quality of response generation in task-oriented dialogue systems. 
DiactTOD addresses the challenge of utilizing generalized dialogue acts to control response generation across different datasets and tasks.
The experimental results on the MultiWOZ dataset show that our approach outperforms previous state-of-the-art methods across a wide range of experimental settings, including zero-shot, few-shot, and full data fine-tuning with both end-to-end and policy optimization configurations.  
Overall, this work demonstrates the effectiveness of DiactTOD, making it possible to build more generalizable end-to-end dialogue systems.

\section*{Limitations}

Despite the effectiveness of our proposed model DiactTOD, we provide some clear limitations.
First, the model is only tested on the MultiWOZ dataset, which is currently the largest dataset for task-oriented response generation. 
While MultiWOZ is a popular dataset in the research community, it is not clear how well the model would perform on other types of datasets or in other domains, particularly those that do not rely on dialogue state annotations. 
It could be an area for future research, by testing the model on other datasets or in other domains to evaluate its robustness and generalizability.

Second, our approach requires a pre-defined dialogue act schema to generate all the possible combinations of dialogue acts. 
This means that it may not be able to generalize well to real-world scenarios where the dialogue acts are not as clearly defined or labeled. 
In those situations, the model may struggle to generate appropriate responses or understand the context. 
In future work, we will develop methods that can adapt to different dialogue act schemas or operate without them.

Another limitation of this work is that the controlled response generation method used is hand-crafted, as opposed to using reinforcement learning. 
We defined rules to control dialogue acts based on the evaluation metrics "Inform" and "Success" of the MultiWOZ dataset. 
This approach may not be suitable for more complex scenarios where the dialogue acts are more varied and thus may require a larger model to build the necessary rules.
Also, Inform and Success metrics may not reflect the real performance and have limitations.
In those situations, alternative methods such as reinforcement learning may be more appropriate.





\bibliography{anthology,custom}

\begin{thebibliography}{36}
\expandafter\ifx\csname natexlab\endcsname\relax\def\natexlab#1{#1}\fi

\bibitem[{Bao et~al.(2020)Bao, He, Wang, Wu, and Wang}]{bao-etal-2020-plato}
Siqi Bao, Huang He, Fan Wang, Hua Wu, and Haifeng Wang. 2020.
\newblock \href {https://doi.org/10.18653/v1/2020.acl-main.9} {{PLATO}:
  Pre-trained dialogue generation model with discrete latent variable}.
\newblock In \emph{Proceedings of the 58th Annual Meeting of the Association
  for Computational Linguistics}, pages 85--96, Online. Association for
  Computational Linguistics.

\bibitem[{Budzianowski et~al.(2018)Budzianowski, Wen, Tseng, Casanueva, Ultes,
  Ramadan, and Gasic}]{DBLP:conf/emnlp/BudzianowskiWTC18}
Pawel Budzianowski, Tsung{-}Hsien Wen, Bo{-}Hsiang Tseng, I{\~{n}}igo
  Casanueva, Stefan Ultes, Osman Ramadan, and Milica Gasic. 2018.
\newblock \href {https://aclanthology.org/D18-1547/} {Multiwoz - {A}
  large-scale multi-domain wizard-of-oz dataset for task-oriented dialogue
  modelling}.
\newblock In \emph{Proceedings of the 2018 Conference on Empirical Methods in
  Natural Language Processing, Brussels, Belgium, October 31 - November 4,
  2018}, pages 5016--5026. Association for Computational Linguistics.

\bibitem[{Chen et~al.(2022)Chen, Gong, Wang, Yao, Qi, Wei, Hu, Zhou, Mao, Chen,
  Cheng, and Duan}]{chen-etal-2022-dialogved}
Wei Chen, Yeyun Gong, Song Wang, Bolun Yao, Weizhen Qi, Zhongyu Wei, Xiaowu Hu,
  Bartuer Zhou, Yi~Mao, Weizhu Chen, Biao Cheng, and Nan Duan. 2022.
\newblock \href {https://doi.org/10.18653/v1/2022.acl-long.333} {Dialogved: A
  pre-trained latent variable encoder-decoder model for dialog response
  generation}.
\newblock In \emph{Proceedings of the 60th Annual Meeting of the Association
  for Computational Linguistics (Volume 1: Long Papers)}, pages 4852--4864,
  Dublin, Ireland. Association for Computational Linguistics.

\bibitem[{Chen et~al.(2013)Chen, Wang, and Rudnicky}]{chen2013dialogue}
Yun-Nung Chen, William Wang, and Alexander Rudnicky. 2013.
\newblock \href {https://doi.org/10.1109/ASRU.2013.6707716} {Unsupervised
  induction and filling of semantic slots for spoken dialogue systems using
  frame-semantic parsing}.
\newblock pages 120--125.

\bibitem[{Cholakov and Kolev(2022)}]{DBLP:journals/corr/abs-2208-07097}
Radostin Cholakov and Todor Kolev. 2022.
\newblock \href {https://doi.org/10.48550/arXiv.2208.07097} {Efficient
  task-oriented dialogue systems with response selection as an auxiliary task}.
\newblock \emph{CoRR}, abs/2208.07097.

\bibitem[{Gu et~al.(2021)Gu, Wu, Wu, Shi, and Yu}]{DBLP:conf/acl/GuWWSY20}
Jing Gu, Qingyang Wu, Chongruo Wu, Weiyan Shi, and Zhou Yu. 2021.
\newblock \href {https://doi.org/10.18653/v1/2021.acl-short.40} {{PRAL:} {A}
  tailored pre-training model for task-oriented dialog generation}.
\newblock In \emph{Proceedings of the 59th Annual Meeting of the Association
  for Computational Linguistics and the 11th International Joint Conference on
  Natural Language Processing, {ACL/IJCNLP} 2021, (Volume 2: Short Papers),
  Virtual Event, August 1-6, 2021}, pages 305--313. Association for
  Computational Linguistics.

\bibitem[{He et~al.(2022)He, Dai, Zheng, Wu, Cao, Liu, Jiang, Yang, Huang, Si
  et~al.}]{he2022galaxy}
Wanwei He, Yinpei Dai, Yinhe Zheng, Yuchuan Wu, Zheng Cao, Dermot Liu, Peng
  Jiang, Min Yang, Fei Huang, Luo Si, et~al. 2022.
\newblock Galaxy: A generative pre-trained model for task-oriented dialog with
  semi-supervised learning and explicit policy injection.
\newblock \emph{Proceedings of the AAAI Conference on Artificial Intelligence}.

\bibitem[{Hosseini-Asl et~al.(2020)Hosseini-Asl, McCann, Wu, Yavuz, and
  Socher}]{HosseiniAsl2020ASL}
Ehsan Hosseini-Asl, Bryan McCann, Chien-Sheng Wu, Semih Yavuz, and Richard
  Socher. 2020.
\newblock A simple language model for task-oriented dialogue.
\newblock \emph{ArXiv}, abs/2005.00796.

\bibitem[{Lee(2021)}]{Lee2021ImprovingET}
Yohan Lee. 2021.
\newblock Improving end-to-end task-oriented dialog system with a simple
  auxiliary task.
\newblock In \emph{EMNLP}.

\bibitem[{Lewis et~al.(2020)Lewis, Liu, Goyal, Ghazvininejad, Mohamed, Levy,
  Stoyanov, and Zettlemoyer}]{DBLP:conf/acl/LewisLGGMLSZ20}
Mike Lewis, Yinhan Liu, Naman Goyal, Marjan Ghazvininejad, Abdelrahman Mohamed,
  Omer Levy, Veselin Stoyanov, and Luke Zettlemoyer. 2020.
\newblock \href {https://doi.org/10.18653/v1/2020.acl-main.703} {{BART:}
  denoising sequence-to-sequence pre-training for natural language generation,
  translation, and comprehension}.
\newblock In \emph{Proceedings of the 58th Annual Meeting of the Association
  for Computational Linguistics, {ACL} 2020, Online, July 5-10, 2020}, pages
  7871--7880. Association for Computational Linguistics.

\bibitem[{Li et~al.(2018)Li, Panda, Liu, and Gao}]{li2018microsoft}
Xiujun Li, Sarah Panda, Jingjing Liu, and Jianfeng Gao. 2018.
\newblock Microsoft dialogue challenge: Building end-to-end task-completion
  dialogue systems.
\newblock \emph{arXiv preprint arXiv:1807.11125}.

\bibitem[{Lubis et~al.(2020)Lubis, Geishauser, Heck, Lin, Moresi, Niekerk, and
  Gasic}]{lubis2020lava}
Nurul Lubis, Christian Geishauser, Michael Heck, Hsien-Chin Lin, Marco Moresi,
  Carel Niekerk, and Milica Gasic. 2020.
\newblock \href {https://doi.org/10.18653/v1/2020.coling-main.41} {Lava: Latent
  action spaces via variational auto-encoding for dialogue policy
  optimization}.
\newblock pages 465--479.

\bibitem[{Mosig et~al.(2020)Mosig, Mehri, and Kober}]{mosig2020star}
Johannes E.~M. Mosig, Shikib Mehri, and Thomas Kober. 2020.
\newblock \href {http://arxiv.org/abs/2010.11853} {{STAR: A Schema-Guided
  Dialog Dataset for Transfer Learning}}.
\newblock \emph{arXiv e-prints}.

\bibitem[{Nekvinda and Dusek(2021)}]{DBLP:journals/corr/abs-2106-05555}
Tom{\'{a}}s Nekvinda and Ondrej Dusek. 2021.
\newblock \href {http://arxiv.org/abs/2106.05555} {Shades of bleu, flavours of
  success: The case of multiwoz}.
\newblock \emph{CoRR}, abs/2106.05555.

\bibitem[{Paolini et~al.(2021{\natexlab{a}})Paolini, Athiwaratkun, Krone, Ma,
  Achille, Anubhai, dos Santos, Xiang, and
  Soatto}]{DBLP:conf/iclr/PaoliniAKMAASXS21}
Giovanni Paolini, Ben Athiwaratkun, Jason Krone, Jie Ma, Alessandro Achille,
  Rishita Anubhai, C{\'{\i}}cero~Nogueira dos Santos, Bing Xiang, and Stefano
  Soatto. 2021{\natexlab{a}}.
\newblock \href {https://openreview.net/forum?id=US-TP-xnXI} {Structured
  prediction as translation between augmented natural languages}.
\newblock In \emph{9th International Conference on Learning Representations,
  {ICLR} 2021, Virtual Event, Austria, May 3-7, 2021}. OpenReview.net.

\bibitem[{Paolini et~al.(2021{\natexlab{b}})Paolini, Athiwaratkun, Krone, Ma,
  Achille, Anubhai, dos Santos, Xiang, and Soatto}]{tanl}
Giovanni Paolini, Ben Athiwaratkun, Jason Krone, Jie Ma, Alessandro Achille,
  Rishita Anubhai, Cicero~Nogueira dos Santos, Bing Xiang, and Stefano Soatto.
  2021{\natexlab{b}}.
\newblock Structured prediction as translation between augmented natural
  languages.
\newblock In \emph{9th International Conference on Learning Representations,
  {ICLR} 2021}.

\bibitem[{Paul et~al.(2019)Paul, Goel, and
  Hakkani{-}T{\"{u}}r}]{DBLP:conf/interspeech/PaulGH19}
Shachi Paul, Rahul Goel, and Dilek Hakkani{-}T{\"{u}}r. 2019.
\newblock \href {https://doi.org/10.21437/Interspeech.2019-1866} {Towards
  universal dialogue act tagging for task-oriented dialogues}.
\newblock In \emph{Interspeech 2019, 20th Annual Conference of the
  International Speech Communication Association, Graz, Austria, 15-19
  September 2019}, pages 1453--1457. {ISCA}.

\bibitem[{Peng et~al.(2022)Peng, Galley, He, Brockett, Lid{\'e}n, Nouri, Yu,
  Dolan, and Gao}]{Peng2022GODELLP}
Baolin Peng, Michel Galley, Pengcheng He, Chris Brockett, Lars Lid{\'e}n, Elnaz
  Nouri, Zhou Yu, Bill Dolan, and Jianfeng Gao. 2022.
\newblock Godel: Large-scale pre-training for goal-directed dialog.
\newblock \emph{ArXiv}, abs/2206.11309.

\bibitem[{Raffel et~al.(2019)Raffel, Shazeer, Roberts, Lee, Narang, Matena,
  Zhou, Li, and Liu}]{DBLP:journals/corr/abs-1910-10683}
Colin Raffel, Noam Shazeer, Adam Roberts, Katherine Lee, Sharan Narang, Michael
  Matena, Yanqi Zhou, Wei Li, and Peter~J. Liu. 2019.
\newblock \href {http://arxiv.org/abs/1910.10683} {Exploring the limits of
  transfer learning with a unified text-to-text transformer}.
\newblock \emph{CoRR}, abs/1910.10683.

\bibitem[{Rastogi et~al.(2020)Rastogi, Zang, Sunkara, Gupta, and
  Khaitan}]{rastogi2020towards}
Abhinav Rastogi, Xiaoxue Zang, Srinivas Sunkara, Raghav Gupta, and Pranav
  Khaitan. 2020.
\newblock Towards scalable multi-domain conversational agents: The
  schema-guided dialogue dataset.
\newblock In \emph{Proceedings of the AAAI Conference on Artificial
  Intelligence}, volume~34, pages 8689--8696.

\bibitem[{Reimers and Gurevych(2019)}]{reimers-2019-sentence-bert}
Nils Reimers and Iryna Gurevych. 2019.
\newblock \href {https://arxiv.org/abs/1908.10084} {Sentence-bert: Sentence
  embeddings using siamese bert-networks}.
\newblock In \emph{Proceedings of the 2019 Conference on Empirical Methods in
  Natural Language Processing}. Association for Computational Linguistics.

\bibitem[{Roller et~al.(2021)Roller, Dinan, Goyal, Ju, Williamson, Liu, Xu,
  Ott, Smith, Boureau, and Weston}]{DBLP:conf/eacl/RollerDGJWLXOSB21}
Stephen Roller, Emily Dinan, Naman Goyal, Da~Ju, Mary Williamson, Yinhan Liu,
  Jing Xu, Myle Ott, Eric~Michael Smith, Y{-}Lan Boureau, and Jason Weston.
  2021.
\newblock \href {https://doi.org/10.18653/v1/2021.eacl-main.24} {Recipes for
  building an open-domain chatbot}.
\newblock In \emph{Proceedings of the 16th Conference of the European Chapter
  of the Association for Computational Linguistics: Main Volume, {EACL} 2021,
  Online, April 19 - 23, 2021}, pages 300--325. Association for Computational
  Linguistics.

\bibitem[{Schulz et~al.(2017)Schulz, Zumer, El~Asri, and
  Sharma}]{schulz-etal-2017-frame}
Hannes Schulz, Jeremie Zumer, Layla El~Asri, and Shikhar Sharma. 2017.
\newblock \href {https://doi.org/10.18653/v1/W17-2626} {A frame tracking model
  for memory-enhanced dialogue systems}.
\newblock In \emph{Proceedings of the 2nd Workshop on Representation Learning
  for {NLP}}, pages 219--227, Vancouver, Canada. Association for Computational
  Linguistics.

\bibitem[{Shalyminov et~al.(2020)Shalyminov, Sordoni, Atkinson, and
  Schulz}]{shalyminov2020fast}
Igor Shalyminov, Alessandro Sordoni, Adam Atkinson, and Hannes Schulz. 2020.
\newblock \href
  {https://www.microsoft.com/en-us/research/publication/fast-domain-adaptation-for-goal-oriented-dialogue-using-a
  -hybrid-generative-retrieval-transformer/} {Fast domain adaptation for
  goal-oriented dialogue using a hybrid generative-retrieval transformer}.
\newblock In \emph{2020 IEEE International Conference on Acoustics, Speech and
  Signal Processing (ICASSP)}.

\bibitem[{Su et~al.(2022)Su, Shu, Mansimov, Gupta, Cai, Lai, and
  Zhang}]{DBLP:conf/acl/SuSMG0LZ22}
Yixuan Su, Lei Shu, Elman Mansimov, Arshit Gupta, Deng Cai, Yi{-}An Lai, and
  Yi~Zhang. 2022.
\newblock \href {https://doi.org/10.18653/v1/2022.acl-long.319} {Multi-task
  pre-training for plug-and-play task-oriented dialogue system}.
\newblock In \emph{Proceedings of the 60th Annual Meeting of the Association
  for Computational Linguistics (Volume 1: Long Papers), {ACL} 2022, Dublin,
  Ireland, May 22-27, 2022}, pages 4661--4676. Association for Computational
  Linguistics.

\bibitem[{Sun et~al.(2022{\natexlab{a}})Sun, Bao, Wu, and
  He}]{DBLP:conf/naacl/Sun0W022}
Haipeng Sun, Junwei Bao, Youzheng Wu, and Xiaodong He. 2022{\natexlab{a}}.
\newblock \href {https://doi.org/10.18653/v1/2022.findings-naacl.166} {{BORT:}
  back and denoising reconstruction for end-to-end task-oriented dialog}.
\newblock In \emph{Findings of the Association for Computational Linguistics:
  {NAACL} 2022, Seattle, WA, United States, July 10-15, 2022}, pages
  2156--2170. Association for Computational Linguistics.

\bibitem[{Sun et~al.(2022{\natexlab{b}})Sun, Bao, Wu, and
  He}]{DBLP:journals/corr/abs-2210-08917}
Haipeng Sun, Junwei Bao, Youzheng Wu, and Xiaodong He. 2022{\natexlab{b}}.
\newblock \href {https://doi.org/10.48550/arXiv.2210.08917} {Mars:
  Semantic-aware contrastive learning for end-to-end task-oriented dialog}.
\newblock \emph{CoRR}, abs/2210.08917.

\bibitem[{Wang et~al.(2020{\natexlab{a}})Wang, Zhang, Kim, and
  Gu}]{DBLP:journals/corr/abs-2006-06814}
Jianhong Wang, Yuan Zhang, Tae{-}Kyun Kim, and Yunjie Gu. 2020{\natexlab{a}}.
\newblock \href {http://arxiv.org/abs/2006.06814} {Modelling hierarchical
  structure between dialogue policy and natural language generator with option
  framework for task-oriented dialogue system}.
\newblock \emph{CoRR}, abs/2006.06814.

\bibitem[{Wang et~al.(2020{\natexlab{b}})Wang, Tian, Wang, Quan, and
  Yu}]{DBLP:conf/acl/WangTWQY20}
Kai Wang, Junfeng Tian, Rui Wang, Xiaojun Quan, and Jianxing Yu.
  2020{\natexlab{b}}.
\newblock \href {https://doi.org/10.18653/v1/2020.acl-main.638} {Multi-domain
  dialogue acts and response co-generation}.
\newblock In \emph{Proceedings of the 58th Annual Meeting of the Association
  for Computational Linguistics, {ACL} 2020, Online, July 5-10, 2020}, pages
  7125--7134. Association for Computational Linguistics.

\bibitem[{Wu et~al.(2023)Wu, Alnuhait, Chen, and
  Yu}]{DBLP:journals/corr/abs-2305-13710}
Qingyang Wu, Deema Alnuhait, Derek Chen, and Zhou Yu. 2023.
\newblock \href {https://doi.org/10.48550/arXiv.2305.13710} {Using textual
  interface to align external knowledge for end-to-end task-oriented dialogue
  systems}.
\newblock \emph{CoRR}, abs/2305.13710.

\bibitem[{Wu et~al.(2021)Wu, Zhang, Li, and Yu}]{DBLP:conf/eacl/WuZLY21}
Qingyang Wu, Yichi Zhang, Yu~Li, and Zhou Yu. 2021.
\newblock \href {https://doi.org/10.18653/v1/2021.eacl-main.110} {Alternating
  recurrent dialog model with large-scale pre-trained language models}.
\newblock In \emph{Proceedings of the 16th Conference of the European Chapter
  of the Association for Computational Linguistics: Main Volume, {EACL} 2021,
  Online, April 19 - 23, 2021}, pages 1292--1301. Association for Computational
  Linguistics.

\bibitem[{Yang et~al.(2021)Yang, Li, and Quan}]{DBLP:conf/aaai/YangLQ21}
Yunyi Yang, Yunhao Li, and Xiaojun Quan. 2021.
\newblock \href {https://ojs.aaai.org/index.php/AAAI/article/view/17674}
  {{UBAR:} towards fully end-to-end task-oriented dialog system with {GPT-2}}.
\newblock In \emph{Thirty-Fifth {AAAI} Conference on Artificial Intelligence,
  {AAAI} 2021, Thirty-Third Conference on Innovative Applications of Artificial
  Intelligence, {IAAI} 2021, The Eleventh Symposium on Educational Advances in
  Artificial Intelligence, {EAAI} 2021, Virtual Event, February 2-9, 2021},
  pages 14230--14238. {AAAI} Press.

\bibitem[{Yu et~al.(2022)Yu, Wu, Qian, and
  Yu}]{https://doi.org/10.48550/arxiv.2211.16773}
Xiao Yu, Qingyang Wu, Kun Qian, and Zhou Yu. 2022.
\newblock \href {https://doi.org/10.48550/ARXIV.2211.16773} {Krls: Improving
  end-to-end response generation in task oriented dialog with reinforced
  keywords learning}.

\bibitem[{Zang et~al.(2020)Zang, Rastogi, Sunkara, Gupta, Zhang, and
  Chen}]{zang-etal-2020-multiwoz}
Xiaoxue Zang, Abhinav Rastogi, Srinivas Sunkara, Raghav Gupta, Jianguo Zhang,
  and Jindong Chen. 2020.
\newblock \href {https://doi.org/10.18653/v1/2020.nlp4convai-1.13}
  {{M}ulti{WOZ} 2.2 : A dialogue dataset with additional annotation corrections
  and state tracking baselines}.
\newblock In \emph{Proceedings of the 2nd Workshop on Natural Language
  Processing for Conversational AI}, pages 109--117, Online. Association for
  Computational Linguistics.

\bibitem[{Zhang et~al.(2020)Zhang, Sun, Galley, Chen, Brockett, Gao, Gao, Liu,
  and Dolan}]{DBLP:conf/acl/ZhangSGCBGGLD20}
Yizhe Zhang, Siqi Sun, Michel Galley, Yen{-}Chun Chen, Chris Brockett, Xiang
  Gao, Jianfeng Gao, Jingjing Liu, and Bill Dolan. 2020.
\newblock \href {https://doi.org/10.18653/v1/2020.acl-demos.30} {{DIALOGPT} :
  Large-scale generative pre-training for conversational response generation}.
\newblock In \emph{Proceedings of the 58th Annual Meeting of the Association
  for Computational Linguistics: System Demonstrations, {ACL} 2020, Online,
  July 5-10, 2020}, pages 270--278. Association for Computational Linguistics.

\bibitem[{Zhao et~al.(2019)Zhao, Xie, and Eskenazi}]{zhao2019rethinking}
Tiancheng Zhao, Kaige Xie, and Maxine Eskenazi. 2019.
\newblock Rethinking action spaces for reinforcement learning in end-to-end
  dialog agents with latent variable models.
\newblock \emph{arXiv preprint arXiv:1902.08858}.

\end{thebibliography}
\bibliographystyle{acl_natbib}

\newpage

\appendix

\section{Dataset Details}

We chose four datasets that are annotated with dialogue acts and one dataset that does not contain any dialogue act annotations. 
Their detailed descriptions are below:

\begin{itemize}
  \item SGD \cite{rastogi2020towards} is a dataset with multi-domain and multi-turn task-oriented conversations between a human and a bot.
  It involves 20 domains including banks, events, media, calendar, travel, and weather.
 
  \item STAR \cite{mosig2020star} is a schema-guided dialogue dataset with human-human conversations across 13 different domains. It designs a flow chart and schema graph for collecting the data.
 
  \item MSRe2e \cite{li2018microsoft} contains 2,890 human-human conversation with three task domains including movie-ticket booking, restaurant reservation, and taxi ordering. 
  
  \item Frames \cite{schulz-etal-2017-frame} is a dataset with 1,369 human-human dialogues. It includes round-trip flights and hotel booking. It uses semantic frames to summarize the dialogue history and states.
  
  \item MetaLWOZ \cite{shalyminov2020fast} is a large dataset containing 37,884 dialogues with domains including bus schedules, apartment search, alarm setting, banking, and event reservation. However, compared to other task-oriented dialogue datasets, this dataset does not provide natural language understanding annotations, and cannot directly be used for end-to-end task-oriented dialogue systems.
  
\end{itemize}

\begin{table*}[h]
    \centering
    \begin{tabular}{l|c c c |c c c c }
    \toprule
    \textbf{Model} & \textbf{Labeled?} & \textbf{Gold Act} & \textbf{S-BERT*} & \textbf{Inform} & \textbf{Success} & \textbf{BLEU} & \textbf{Combined} \\
    \midrule
    $\text{LaDiact}_\text{Base}$ & Unlabeled & no & no & 66.1 & 30.6 & 1.43 & 49.78 \\
     & Unlabeled & yes & no & 94.0 & 42.2 & 0.17 & 68.27 \\
     & Unlabeled & no & yes & 89.4 & 51.0 & 0.59 & 70.79 \\
     & Unlabeled & yes & yes & 78.8 & 41.9 & 0.87 & 61.22 \\
    \midrule
    $\text{LaDiact}_\text{Base}$ & Labeled & no & no & 84.7 & 46.3 & 4.11 & 69.61 \\
     & Labeled & yes & no & 83.8 & 47.0 & 5.21 & 70.62 \\
     & Labeled & no & yes & 84.8 & 47.0 & 4.59 & 70.49 \\
     & Labeled & yes & yes & 91.3 & 51.9 & 6.05 & 77.65 \\
    \midrule
    $\text{LaDiact}_\text{Base}$ & Mixed & no & no & 84.6 & 47.5 & 4.49 & 70.54 \\
     & Mixed & yes & no & 94.3 & 54.3 & 6.62 & 80.92 \\
     & Mixed & no & yes & 87.8 & 50.6 & 5.18 & 74.38 \\
     & Mixed & yes & yes & 93.2 & 54.6 & 6.56 & 80.46 \\
    \midrule
    \midrule
    $\text{LaDiact}_\text{Large}$ & Unlabeled & no & no & 72.0 & 30.6 & 1.44 & 52.74 \\
     & Unlabeled & yes & no & 81.7 & 46.2 & 0.17 & 63.95 \\
     & Unlabeled & no & yes & 68.0 & 38.8 & 3.73 & 57.13 \\
     & Unlabeled & yes & yes & 79.7 & 42.4 & 2.03 & 63.80 \\
    \midrule
    $\text{LaDiact}_\text{Large}$ & Labeled & no & no & 87.8 & 49.1 & 4.89 & 73.33 \\
     & Labeled & yes & no & 93.3 & 48.9 & 6.40 & 77.50 \\
     & Labeled & no & yes & 93.1 & 53.9 & 4.79 & 78.29 \\
     & Labeled & yes & yes & 92.5 & 53.7 & 6.25 & 79.35 \\
    \midrule
    $\text{LaDiact}_\text{Large}$ & Mixed & no & no & 90.5 & 52.8 & 5.11 & 76.76 \\
     & Mixed & yes & no & 92.2 & 55.5 & 6.67 & 80.52 \\
     & Mixed & no & yes & 91.4 & 53.0 & 5.05 & 77.25 \\
     & Mixed & yes & yes & 93.8 & 55.4 & 6.57 & 81.17 \\
    \bottomrule
    \end{tabular}
    \caption{Detailed ablation studies of zero-shot performance under different configurations.. * means whether freezes sentence-BERT during pre-training.}
    \label{tab:result}
\end{table*}

\end{document}